\documentclass[10pt,twocolumn,letterpaper]{article}

\usepackage{iccv}
\usepackage{times}
\usepackage{epsfig}
\usepackage{graphicx}
\usepackage{amsmath}
\usepackage{amssymb}
\usepackage{pifont}
\usepackage{authblk}
\usepackage[T1]{fontenc} 

\usepackage[pagebackref=true,breaklinks=true,letterpaper=true,colorlinks,bookmarks=false]{hyperref}

\iccvfinalcopy 

\def\iccvPaperID{9} 

\ificcvfinal\fi

\author[1,2]{Leon Mlodzian}
\author[2]{Zhigang Sun}
\author[3,5]{Hendrik Berkemeyer}
\author[2,3]{Sebastian Monka}
\author[2,4]{Zixu Wang}
\author[1]{Stefan Dietze}
\author[2,3]{Lavdim Halilaj}
\author[2,3]{Juergen Luettin}
\affil[1]{Heinrich Heine University D\"usseldorf}
\affil[2]{Bosch Center for Artificial Intelligence}
\affil[3]{Robert Bosch GmbH}
\affil[4]{Technical University of Munich}
\affil[5]{University of Osnabr\"uck}
\affil[ ]{\tt\small {leon.mlodzian@proton.me, zhigang.sun3@cn.bosch.com, \{hendrik.berkemeyer, sebastian.monka, lavdim.halilaj, juergen.luettin\}@de.bosch.com}, zixu.wang@tum.de, stefan.dietze@hhu.de}

\begin{document}

\title{nuScenes Knowledge Graph - A comprehensive semantic representation of traffic scenes for trajectory prediction}

\maketitle
\ificcvfinal\thispagestyle{empty}\fi

\begin{abstract}

Trajectory prediction in traffic scenes involves accurately forecasting the behaviour of surrounding vehicles. 
To achieve this objective it is crucial to consider contextual information, including the driving path of vehicles, road topology, lane dividers, and traffic rules. 
Although studies demonstrated the potential of leveraging heterogeneous context for improving trajectory prediction, state-of-the-art deep learning approaches still rely on a limited subset of this information. This is mainly due to the limited availability of comprehensive representations.
This paper presents an approach that utilizes knowledge graphs to model the diverse entities and their semantic connections within traffic scenes.
Further, we present nuScenes Knowledge Graph (nSKG), a knowledge graph for the nuScenes dataset, that models explicitly all scene participants and road elements, as well as their semantic and spatial relationships. 
To facilitate the usage of the nSKG via graph neural networks for trajectory prediction, we provide the data in a format, ready-to-use by the $PyG$ library.
All artefacts can be found here:  \url{https://github.com/boschresearch/nuScenes_Knowledge_Graph}.

\end{abstract}
\vspace{-0,5cm}

\section{Introduction}
\label{introduction}

Traffic trajectory prediction is a crucial component of autonomous driving, as it enables the autonomous vehicle to anticipate the movement of other traffic participants and avoid dangerous situations that could lead to collisions. 
Deep learning approaches have proven to be very successful when applied to this task. 
A key driver behind the significant progress in deep learning is the availability of easily accessible datasets that have been compiled over the years. 
For instance, MNIST~\cite{LeCun1998GradientbasedLA}, COCO ~\cite{Lin2014MicrosoftCC} and ImageNet~\cite{5206848} were crucial for progress in computer vision, GLUE~\cite{Wang2018GLUEAM} and SQuAD~\cite{Rajpurkar2016SQuAD1Q} for natural language understanding and MuJoCo~\cite{todorov2012mujoco} and OpenAI Gym~\cite{1606.01540} for reinforcement learning. 
The same is valid for trajectory prediction and includes datasets, such as Argoverse~\cite{Chang2019Argoverse}, 
Apolloscape~\cite{ma2019trafficpredict},
Interaction~\cite{Zhan2019Interaction}, and nuScenes~\cite{Caesar2020nuscenes}.
However, there are two shortcomings of current approaches in trajectory prediction: (1) shortcomings of deep learning and (2) shortcomings in rich scene representation. We will describe them in more detail in the following sections.

\begin{figure}[t]
\begin{center}
   \includegraphics[width=1.0\linewidth]{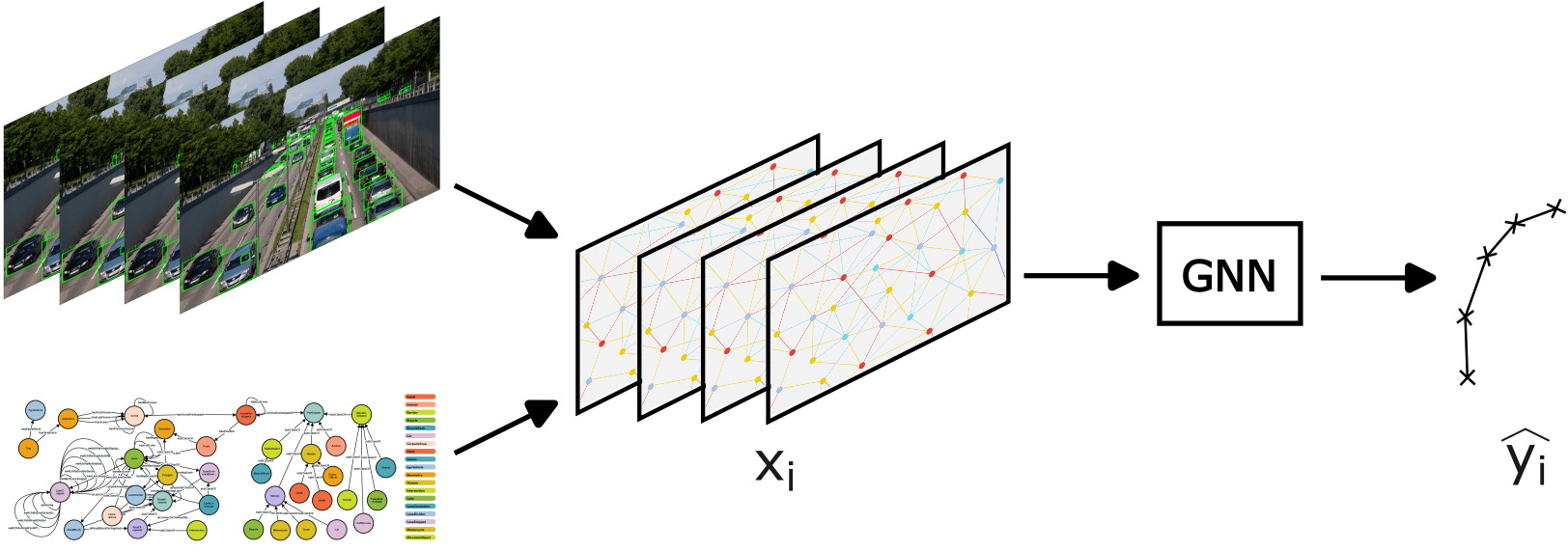}
\end{center}
\vspace{-0,2cm}
   \caption{
   We model traffic scenes (top left) by applying a rigorous ontology (bottom left) to them, producing rich, temporal, heterogeneous graphs. We provide a large graph regression dataset of $(x_i, y_i)$ pairs for training GNNs on the designed representation. Partial image credits: \href{https://www.freepik.com/free-vector/global-communication-background-business-network-vector-design_19585255.htm}  
   {rawpixel.com on Freepik}.
   }
\vspace{-0,5cm}
\end{figure}

\textbf{Shortcomings of deep learning} have been the subject of several investigations over the last years.
Specifically, their lack of robustness~\cite{Szegedy2013IntriguingProperties, Marcus2018DeepLA}, explainability~\cite{Ji2020ASO, Hogan2020KnowledgeG, Lipton2016Mythos} as well as the inability to generalise to new domains~\cite{Zhang2016UnderstandingDL,Battaglia2018RelationalIB}~\cite{Bahari2021VehicleTP}.
One possible explanation for these limitations is that they operate purely on a sub-symbolic~\cite{Monka2022Visual} and statistical basis, thus only learning correlations between input features and target variable, rather than attaining a causal, structured comprehension of a task ~\cite{Scherrer2022OnTG, Ahuja2022InterventionalCR, Scholkopf2021TowardCR}. 
Furthermore, for real-world applications of autonomous systems it is vital to consider safety aspects. 
ISO 26262~\cite{ISO26262}, the international standard for the functional safety of road vehicles, needs to be satisfied. 
Challenges in validation when using machine learning methods have been described in~\cite{Burton2017MakingTC}.

Studies suggest that humans do not reason at the pixel level but use attention and expectation at the object level~\cite{Summerfield2009ExpectationA, denOuden2012HowPE} to do predictive coding~\cite{Clark2013WhateverNP,Butz2021EventPredictiveCA}.
Moreover, we possess inherent prior knowledge, such as intuitive physics and common sense~\cite{Presmeg1992TheBI} that we use in tasks like trajectory prediction.
This high-level, structured information (knowledge) is typically missing when deep learning models are trained in end-to-end scenarios from raw data. 
We address this shortcoming by providing a semantic representation of the driving scene that can be exploited by deep learning based approaches.


\newcommand{\xmark}{{\color{red}\ding{55}}}
\newcommand{\cmark}{{\color{green}\ding{51}}}

\begin{table*}[t]
    \centering
    \begin{tabular}{c|c|c|c|c|c|c|c|c|c|c|c}
         &  Lane    & Lane  & Lane & Border & Stop & Traffic & Traffic & Cross- & Walk- & Car & Agent\\
         &  center & width & border & type & area & light & signs & ing & way  & park & relations \\
       \hline
       VectorNet \cite{Gao2020VectorNetEH} & \xmark & \xmark & \cmark & \xmark & \xmark & \xmark & \cmark & \cmark & \xmark & \xmark & \xmark \\
       LaneGCN \cite{Liang2020LaneGCN} & \cmark & \xmark & \xmark & \xmark & \xmark & \xmark & \xmark & \xmark & \xmark & \xmark & \xmark\\
       Holistic  \cite{Grimm2023HolisticGM} & \cmark & \xmark & \xmark & \xmark & \xmark & \xmark & \xmark & \xmark & \xmark & \xmark & \xmark\\
       Relation \cite{Zipfl2022SceneGraphs} & \cmark & \xmark & \xmark & \xmark & \xmark & \xmark & \xmark & \xmark & \xmark & \xmark & \xmark\\
       PGP \cite{Deo2021MultimodalTP} & \cmark & \xmark & \xmark & \xmark & \cmark & \xmark & \xmark & \cmark & \xmark & \xmark & \xmark\\
       HDGT \cite{Jia2022HDGTHD} & \cmark & \xmark & \xmark & \xmark & \xmark & \cmark & \xmark & \xmark & \xmark & \xmark & \xmark\\
       LAformer \cite{liu2023laformer} & \cmark & \xmark & \xmark & \xmark & \cmark & \xmark & \xmark & \cmark & \xmark & \xmark & \xmark\\
        \textbf{Ours} & \cmark & \cmark & \cmark & \cmark & \cmark & \cmark & \cmark & \cmark & \cmark & \cmark & \cmark
    \end{tabular}
    \caption{Comparison of information included in popular and state-of-the-art trajectory prediction approaches. Raster-based methods \cite{Cui2018MultimodalTP, PhanMinh2019CoverNetMB} were not included due to their inferior performance and non-explicit information structure.
    }
    \label{tab:comparison}
\end{table*}


\textbf{Shortcomings in rich scene representation} describes the situation that trajectory prediction datasets described above, lack in rich scene representation. Especially map and scene context information is rarely included.
nuScenes is a unique dataset for trajectory prediction that stands out due to its comprehensive map information. 
However, the trajectory prediction community has not fully exploited the detailed heterogeneous map data because it is not provided in an easy to use data representation.
Knowledge graphs~\cite{Hogan2020KnowledgeG}, on the other hand, are well suited to represent and reason over structured and high-level information.

\textbf{In this work}, we provide a solution to address both shortcomings, deep learning and rich scene representation.
We leverage the power of knowledge graphs to provide a comprehensive representation of the driving scene, forming a graph-based, symbolic representation at an intermediate level of abstraction.
We implement our approach for the nuScenes dataset and provide the nuScenes Knowledge Graph (nSKG), a comprehensive, semantic representation of driving scenes. 
nSKG utilizes subject-predicate-object triples to structure high-level information. It is based on a rigorous ontology to model concepts such as agents (traffic participants) and map, their hierarchies and relationships.
It is a rich representation of carpark areas, walkways, pedestrian crossings, lane geometry, and other map elements as well as traffic participants, their trajectories and semantic relations, including spatio-temporal relations between entities.
Furthermore, we extract a nuScenes trajectory prediction graph dataset (nSTP) to alleviate data engineering efforts for neural network designers.
It includes the wealth of relevant information from the knowledge graph and thus forms a new scene graph dataset that enables training graph neural networks (GNNs) on our rich scene representation.
Both resources together enable symbolic (nSKG) and sub-symbolic (nSTP) methods to be explored for trajectory prediction with a wealth of structured information, previously only available in unstructured form.
Neuro-symbolic AI has been dubbed the third wave of AI \cite{Garcez2020NeurosymbolicAI} based on the conjecture that the fusion of symbolic and sub-symbolic methods could relieve intelligent systems from the disadvantages of each. 
This could help to obtain explainable models that meet the safety requirements of autonomous vehicles.

The \textbf{main contributions} are:
\begin{itemize}
    \item A comprehensive agent and map ontology that models driving scenes in detail.
    \item nSKG, a knowledge graph generated for the nuScenes dataset, based on the defined ontology.
    \item nSTP, a ready-to-use scene graph dataset for training GNNs for trajectory prediction. 
\end{itemize}

The next section summarises the related work. 
Section~\ref{sec:ontology} presents our ontology design for modeling traffic scenes as well as the generation of the nuScenes Knowledge Graph. 
Section~\ref{sec:dataset} describes the construction of our readily usable graph dataset for trajectory prediction. Section \ref{sec:limitations} states limitations of our work and conclusions follow in the final section.


\section{Related work}\label{related work}

\subsection{Trajectory prediction}

One of the first set of neural networks applied to trajectory prediction were raster-based approaches \cite{Cui2018MultimodalTP,Djuric2018UncertaintyawareSM,PhanMinh2019CoverNetMB,Berkemeyer2021FeasibleAA}. These approaches encode the traffic scene into birds-eye-view images with a number of channels. The channels are used to represent the various kinds of structures and agents in a scene. 
On top of these raster-representations, convolutional neural networks \cite{LeCun1995Convolutional} are applied to learn a representation of the map and agents. 
Drawback of these models is that they do not have access to high-level information and need to learn from raw pixels. 

The next generation of trajectory prediction techniques used a more natural and powerful data representation approach: graphs \cite{Li2019GRIPGI, Li2020EvolveGraphMT, Gao2020VectorNetEH,Liang2020LaneGCN, Li2021GRINGR, Varadarajan2021MultiPathEI, Grimm2023HolisticGM,liu2023laformer}. 
These are higher-level data representations that do not require networks to learn from low-level pixels, which yields performance improvements. 
State-of-the-art approaches use these graphs for data representation. The various methods model scenes at different levels of abstraction. Methods like VectorNet \cite{Gao2020VectorNetEH} use fine representations, where nodes are simply coordinates and in combination with edges between them, they represent map structure borders or vehicle trajectories. 
On the other hand, very recent approaches \cite{Grimm2023HolisticGM, Zipfl2022SceneGraphs} use high-level representations, where single nodes represent whole entities, like vehicles or lanes. For such high-level representations, heterogeneous graphs are employed to capture the different types of nodes and edges that arise.

Graph neural networks are the standard method for learning on graphs. Heterogeneous graphs are either used in conjunction with a heterogeneous graph neural network \cite{Grimm2023HolisticGM}, or the types of nodes and edges is categorically encoded into a feature and then processed by a standard (homogeneous) graph neural network \cite{Gao2020VectorNetEH, Zipfl2022SceneGraphs}.

Traffic representations designed for trajectory prediction have become more structured and high-level over time. From rasters to simple graphs, from simple graphs to heterogeneous graphs and this work takes another step, namely knowledge graphs.

\subsection{Map representation}

Recently, rich map context has received increased attention and is considered to be an important cornerstone in reaching further improvements in trajectory prediction \cite{Liang2020LaneGCN}. 
It is an open research question how the complex and rich road topology with lanes, walkways, car parks, traffic signs, pedestrian crossings and traffic lights is best represented and how much this aids trajectory prediction. 
No previous work has been found that uses available map information comprehensively (see table \ref{tab:comparison}). 
Although being widely considered important \cite{Liang2020LaneGCN, Gao2020VectorNetEH, Deo2021MultimodalTP}, the large majority of map information has so far been ignored, possibly due to the high engineering effort to obtain an easily usable data representation. 
State-of-the-art results in trajectory prediction were reached by \cite{Deo2021MultimodalTP} which includes lane center points, pedestrian crossings and stop area information. We hypothesise that results can be improved by representing more diverse road elements and semantic relational information.

Looking beyond trajectory prediction, there are other branches of automated driving interested in how maps can be represented. A recent survey on knowledge graphs for automated driving \cite{Luettin2022Survey} contains a comprehensive list of available ontologies, only one of which has a focus on the map \cite{Suryawanshi2019AnOM}. It is a small ontology with only seven concepts that explores the feasibility of using ontologies for driver assistance functions. 
The map structures that it models are lanes, traffic signs and road pieces. 
On the other extreme, \cite{Westhofen2022Criticality} uses description logic reasoning to recognise the criticality of driving situations. Things like whether the road is wet or sandy, what a traffic light's color is and much else is considered. The ontology is very complex with a large number of concepts, the large majority of which cannot be generated from trajectory prediction datasets since such information is neither directly included nor derivable.

\subsection{Trajectory representation}

For modelling the trajectories of participants, relevant schemas exist. \cite{Halilaj2022KnowledgeGF} and \cite{Hu2013Geo-ontology} both propose an ontology for modelling agent data. The main difference between them is that the former is agent-centric whereas the latter is trajectory-centric. The agent-centric model includes a notion of agents at certain timesteps. It only models agents as time-independent. In trajectory prediction, one cares about how properties of agents like speed and orientation evolve, making an agent-centric model suitable.

\subsection{Ontologies in autonomous driving}

Some well-known general ontologies that contain concepts related to autonomous driving (AD) include SOSA \cite{Janowicz2018SOSAAL}, DBpedia \cite{Auer2007DBpediaAN} and Schema.org \cite{Guha2015SchemaorgEO}. 
A survey that compares and contrasts available ontologies in AD can be found in~\cite{Luettin2022Survey}.
More specific works include \cite{Geyer2014ConceptAD, Ulbrich2015DefiningAS} that intend to create a shared vocabulary across AD applications. There exist ontologies that model vehicles \cite{Zhao2015CoreOF} and sensors \cite{Klotz2018VSSoTV}. Human driver modelling has received attention in \cite{Hina2017OntologicalAM, Feld2011TheAO} and particularly in \cite{Sarwar2019ContextAO} where demographic and behavioural aspects are considered. A context model for automated vehicles is presented in \cite{Ulbrich2014GraphbasedCR}. This models some of the aspects we are interested in, but many relevant factors for trajectory prediction are not modelled. Lastly, there are standardisation efforts for modelling the central concepts in AD with ontologies, for example ASAM OpenX \cite{ASAMOpenX} and ASAM OpenScenario \cite{ASAMOpenScenario}.
So far, these ontologies have been hardly used due the lack of available data.
Here, we design an ontology to be applied for AD that represents data that is typically expected to be available in future AD systems. 
As first implementation, we choose to use the nuScenes dataset, one of the most widely used datasets in autonomous driving that contains rich map information and that was recorded by a stat-of-the-art sensor suite of Lidar, Radar cameras, IMU and GPS sensor.

\section{Ontology and knowledge graph generation}\label{sec:ontology}

\begin{figure}[t]
\begin{center}
   \includegraphics[width=0.8\linewidth]{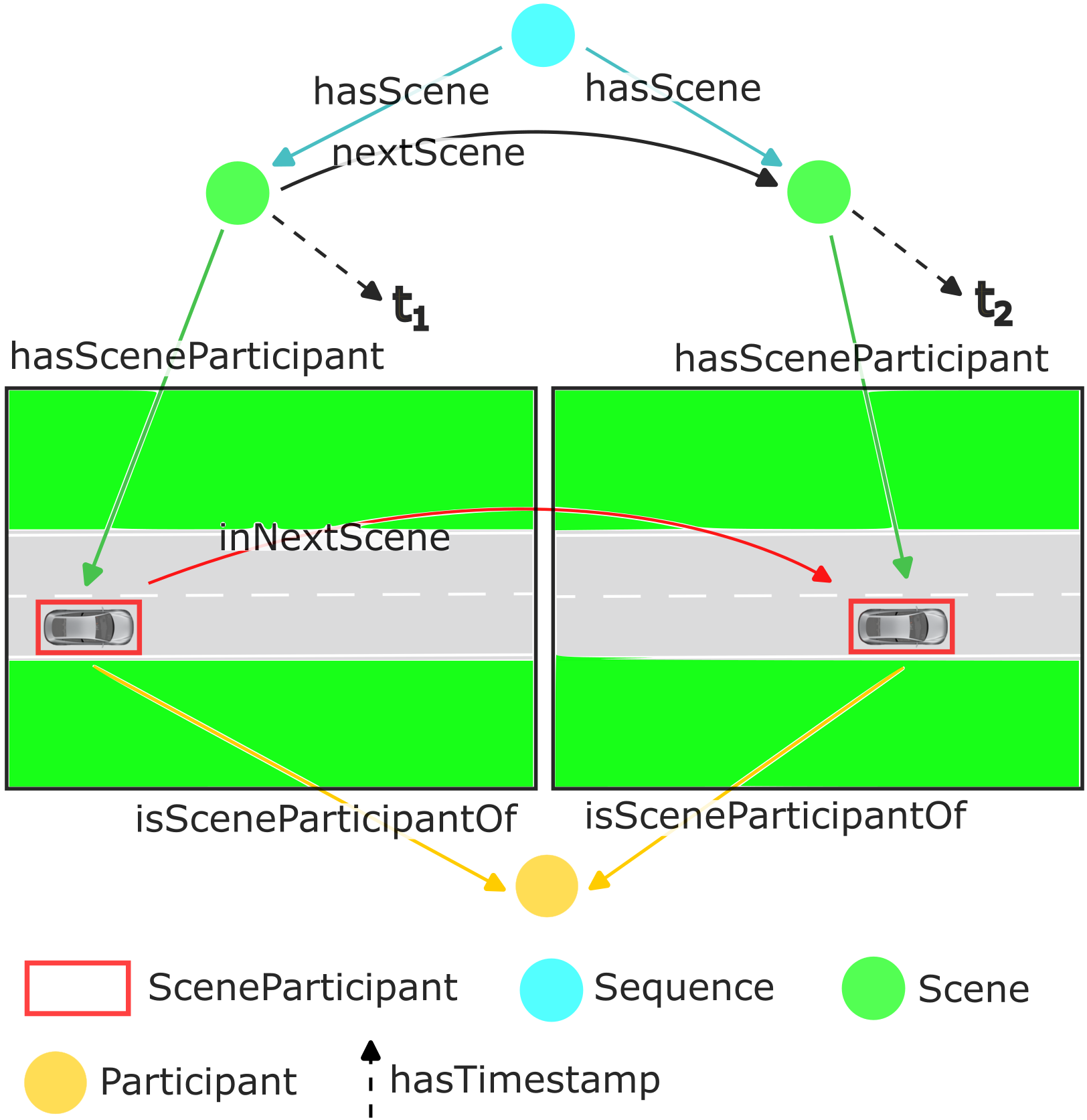}
\end{center}
   \caption{Model of the temporal nature of traffic scenarios applied to a single car travelling along a lane.
   }
\label{fig:temporal}
\end{figure}
To describe the design of the ontology and the generation of the knowledge graph, we first introduce a concept, then a $\mathcal{SROIQ(D)}$ (on which OWL 2 \cite{owl2} is based) description logic formalisation is given, and finally the knowledge graph instance generation from nuScenes is explained.
The  ontology is a generic traffic scene model that can be applied to other datasets or extended to represent new pieces of scene information in future.

\subsection{Temporal representation}

\textbf{Sequence, Scene.} We reuse the concepts \textit{Sequence} and \textit{Scene} from \cite{Halilaj2022KnowledgeGF} to divide a driving situation. A \textit{Scene} refers to a single moment in a traffic situation. \textit{hasTimestamp} is an integer data property with a unix timestamp defining the moment in time. A \textit{Sequence} is an ordered collection of \textit{Scene}s. A \textit{Sequence} can be thought of as a video where its frames are \textit{Scene}s. Since the order of \textit{Scene}s is inherent in them, object properties \textit{hasNextScene} and \textit{hasPreviousScene} are defined to link consecutive \textit{Scene}s.
\begin{equation}
\begin{split}
    \mathit{Scene} \equiv \  & \exists \mathit{hasNextScene}.\mathit{Scene} \\ \cup \  & \exists \mathit{hasPreviousScene}.\mathit{Scene}
\end{split}
\end{equation}
\begin{equation}
    \mathit{Sequence} \equiv \exists \mathit{hasScene}.\mathit{Scene}
\end{equation}

\textit{Sequence} and \textit{Scene} instances are generated from the \textsc{scene} and \textsc{sample} nuScenes records, respectively.

\textbf{Trip, Location.} \textit{Sequence}s refer to specific trajectory prediction situations. During recording of motion data the ego-vehicle might travel for hours and record several \textit{Sequences}. A \textit{Trip} is such a recording session and each of its entities points to several \textit{Sequence}s. Each \textit{Trip} is taken in a particular region of interest, a \textit{Location}, related to it via \textit{hasLocation}. \textit{hasRightHandTraffic} is a boolean property to describe the driving direction at a \textit{Location}.

\begin{equation}
    \mathit{Trip} \equiv \exists \mathit{hasSequence}.\mathit{Sequence}
\end{equation}
\vspace{-0.3cm}
\begin{equation}
    \mathit{Location} \equiv \exists \mathit{hasLocation}^{-1}.\mathit{Trip}
\end{equation}

\textit{Trip} instances are generated from nuScenes \textsc{log} records and a \textit{Location} instance is manually created for each of the four maps.

\subsection{Participant representation}

\begin{figure}[t]
\begin{center}
   \includegraphics[width=0.8\linewidth]{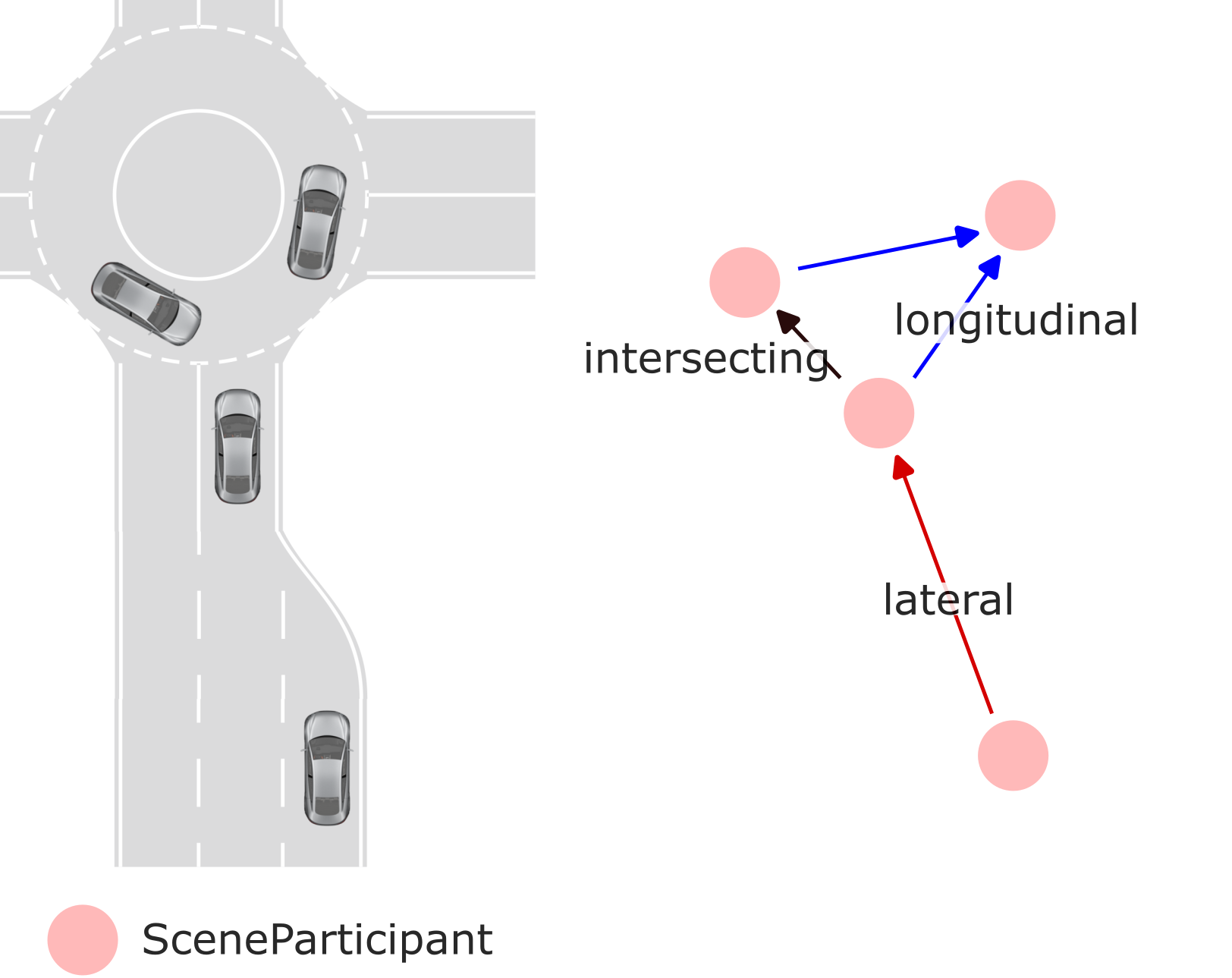}
\end{center}
   \caption{Semantic relationship model between agents.}
\label{fig:agent-agent}
\end{figure}

\textbf{Participant, SceneParticipant.} The \textit{Participant} concept represents a traffic agent present in one or multiple \textit{Scene}s. The various types of participants are modelled as subclasses of the \textit{Participant} concept. There are in total 23 different ones. Examples are cars, adults, children, police officers, ambulances, bicycles and so on. In \cite{Halilaj2022KnowledgeGF}, \textit{Participant}s refer to an entity at a certain timestep. A new relation \textit{inNextScene} was introduced to be able to link entities across time. Further, the concept \textit{SceneParticipant} was introduced as a notion of an agent at a certain timestep and the meaning of \textit{Participant} was changed to represent an agent generally, independent of time. This avoids having to store time-independent information, e.g. sizes of agents, redundantly. The semantic relationship between \textit{SceneParticipant}s is modelled as in \cite{Zipfl2022SceneGraphs}, where agents may follow one another (longitudinal), potentially intersect (intersecting) or be parallel (lateral) to one another (see figure \ref{fig:agent-agent}).
\begin{equation}
\begin{split}
    & \mathit{SceneParticipant} \equiv \\
    & \quad \exists \mathit{hasSceneParticipant}^{-1} . \mathit{Scene} \  \cap \\ 
    & \quad \exists \mathit{isSceneParticipantOf} . \mathit{Participant}
\end{split}
\end{equation}
\begin{equation}
\begin{split}
    & \mathit{Participant} \equiv \\ 
    & \quad \exists \mathit{isSceneParticipantOf}^{-1}.\mathit{SceneParticipant}
\end{split}
\end{equation}

\textit{SceneParticipant} instances are generated from \textsc{sample\_annotation} and \textsc{ego\_pose} records. The \textsc{ego\_pose} records are needed such that the ego-vehicle can be included as a \textit{SceneParticipant}. This is a novelty in our data representation. Previous work has ignored the effect of the ego-vehicle on the target vehicle's motion. Our data analysis of the nuScenes dataset shows that ego and target can be up to 2m close in a significant number of cases. We therefore expect the ego-vehicle to have an influence on the target vehicle's behaviour.

\subsection{Lane representation}

\textbf{Lane, LaneConnector.} The central component of road traffic infrastructure is the \textit{Lane}. This is defined as a non-overlapping stretch of road surface, typically confined by lane borders, where only one driving direction is allowed. This is a physical lane formalisation as opposed to a logical one, where lanes go across junctions and can overlap \cite{Poggenhans2018Lanelet2}. To keep the logical connectivity information with the physical definition, one needs \textit{LaneConnector}s, which have the functional properties \textit{hasIncomingLane} and \textit{hasOutgoingLane} pointing to a \textit{Lane} each.
\begin{equation}
\begin{split}
    \mathit{Lane} \equiv \  & \exists \mathit{hasNextLane} . \mathit{Lane} \\
    \cup \  & \exists \mathit{hasPreviousLane} . \mathit{Lane} \\
    \cup \  & \exists \mathit{hasLeftLane} . \mathit{Lane} \\
    \cup \  & \exists \mathit{hasRightLane} . \mathit{Lane} 
\end{split}
\end{equation}
\begin{equation}
\begin{split}
    \mathit{LaneConnector} \equiv \  & \exists \mathit{hasIncomingLane}.\mathit{Lane} \\
    \cap \  & \exists \mathit{hasOutgoingLane}.\mathit{Lane}
\end{split}
\end{equation}

\textit{Lane} and \textit{LaneConnector} instances are generated from \textsc{lane} and \textsc{lane\_connector} records, respectively.

\textbf{LaneSnippet, switchVia.} Lane borders are another crucial element determining how cars travel. Different lane divider types exist, such as solid lines and dashed lines. A \textit{LaneSnippet} is defined as a piece of a lane that has a single border type on each its left and its right side. This allows the introduction of a \textit{switchVia} property for every type of border, i.e. \textit{switchViaDoubleDashed}, \textit{switchViaSingleSolid}, etc. Neighbouring snippets that have a , say, single solid border between them, get related to one another via \textit{switchViaSingleSolid}, representing that a single solid border would have to be crossed to switch from one to the other. Switches via borders that are illegal are kept in the model because cars may sometimes break traffic rules and overtake across a solid border, for example. \textit{hasNextLaneSnippet} points from one snippet to the immediately following one and \textit{hasLaneSnippet} keeps them connected to their parent \textit{Lane}. Further, since experimental evidence \cite{Deo2021MultimodalTP} has shown that it is important for trajectory prediction performance to keep snippets short, they are further divided if they exceed 20 meters in length. \textit{snippetHasLength} keeps a record of how long a particular lane snippet is.
\begin{equation}
\begin{split}
    \mathit{LaneSnippet} \equiv \  & \exists \mathit{switchViaDoubleDashed}.\mathit{LaneSnippet} \\
    \cup \  & \exists \mathit{switchViaSingleSolid}.\mathit{LaneSnippet} \\
    \cup \  & \dots
\end{split}
\end{equation}

\textit{LaneSnippet} instances were computed from \textsc{lane} records. 
The border types (solid line, dashed line, etc.) on each side of a lane were tracked and split into sections that have non-changing border types on either side. 
Sections were divided, if necessary, to satisfy the 20m length bound. 
This produced \textit{LaneSnippet} instances with constant border types on either side. 
\textit{switchVia} edges were placed between neighbouring \textit{LaneSnippet} instances.

\textbf{LaneSlice, OrderedPose.} To represent the centerlines (where cars typically drive) of lanes and lane connectors, a sequence of \textit{Pose}s is used. 
A \textit{Pose} consists of a position and an orientation. The orientation here denotes the orientation of the lane, i.e. the traffic direction, at a certain position. 
An \textit{OrderedPose} is a subclass of \textit{Pose} that also has the \textit{hasNextPose} property. This is used to order them, defining the typical trajectory along a \textit{LaneConnector} via the \textit{connectorHasPose} relation to all its ordered poses. 
A \textit{Pose}'s position, is modelled with \textit{sf:Point} as are the agent positions, and its orientation with data property \textit{poseHasOrientation}, represented as the angle between the positive x-axis and the direction facing (yaw). 
Contrary to lane connectors, the lane model needs to satisfy competency questions about width, too. 
The natural naming \textit{LaneSlice} is chosen to represent the combination of center pose and lane width. 
\textit{hasNextLaneSlice} keeps them ordered by connecting consecutive slices, \textit{hasLaneSlice} points from parent lane to its slices and \textit{laneSliceHasWidth} is the data property the name suggests.
\begin{equation}
\begin{split}
    \mathit{LaneSlice} \equiv \  & \exists \mathit{laneHasSlice}^{-1}.\mathit{Lane} \\
    \cap \  & \exists \mathit{laneSliceHasWidth}. \mathbb{R}
\end{split}
\end{equation}
\begin{equation}
\begin{split}
    \mathit{OrderedPose} \equiv \exists & \mathit{connectorHasPose}^{-1} . \\ 
    & \mathit{LaneConnector}
\end{split}
\end{equation}
\begin{equation}
    \mathit{OrderedPose} \sqsubseteq \mathit{Pose}
\end{equation}

\textit{OrderedPose} instances were generated from the hand-annotated arclines from nuScenes at a resolution of 2m with the aid of the nuscenes-devkit. \textit{LaneSlice} instances additionally represent lane width. A given center point, for
which width is to be computed, is projected to both left and right borders. The projected points are those points on the borders that have the smallest Euclidean distance to the given center point. The width is given by the distance between the projected points.

\subsection{Road infrastructure representation}

\textbf{StopArea.} Stop areas are a very important concept for trajectory prediction because they, by definition, are the regions where cars tend to come to a halt. Several reasons exist for such regions and each is modelled as a subclass of the parent class \textit{StopArea}. 
These include stop signs, yield signs, oncoming traffic when wanting to make a left turn, pedestrian crossings and traffic lights. \textit{causesStopAt} link the causing entity to their associated \textit{StopArea}.
\begin{equation}
\begin{split}
    & \mathit{StopArea} \equiv \mathit{PedCrossingStopArea} \\ 
    & \quad \cup \  \mathit{TrafficLightStopArea}  \cup  \mathit{YieldStopArea} \\ 
    & \quad \cup \  \mathit{StopSignArea}  \cup  \mathit{TurnStopArea}
\end{split}
\end{equation}

\textit{StopArea} instances were generated from nuScenes \textsc{stop\_line} records.

\begin{figure}[t]
\begin{center}
   \includegraphics[width=0.8\linewidth]{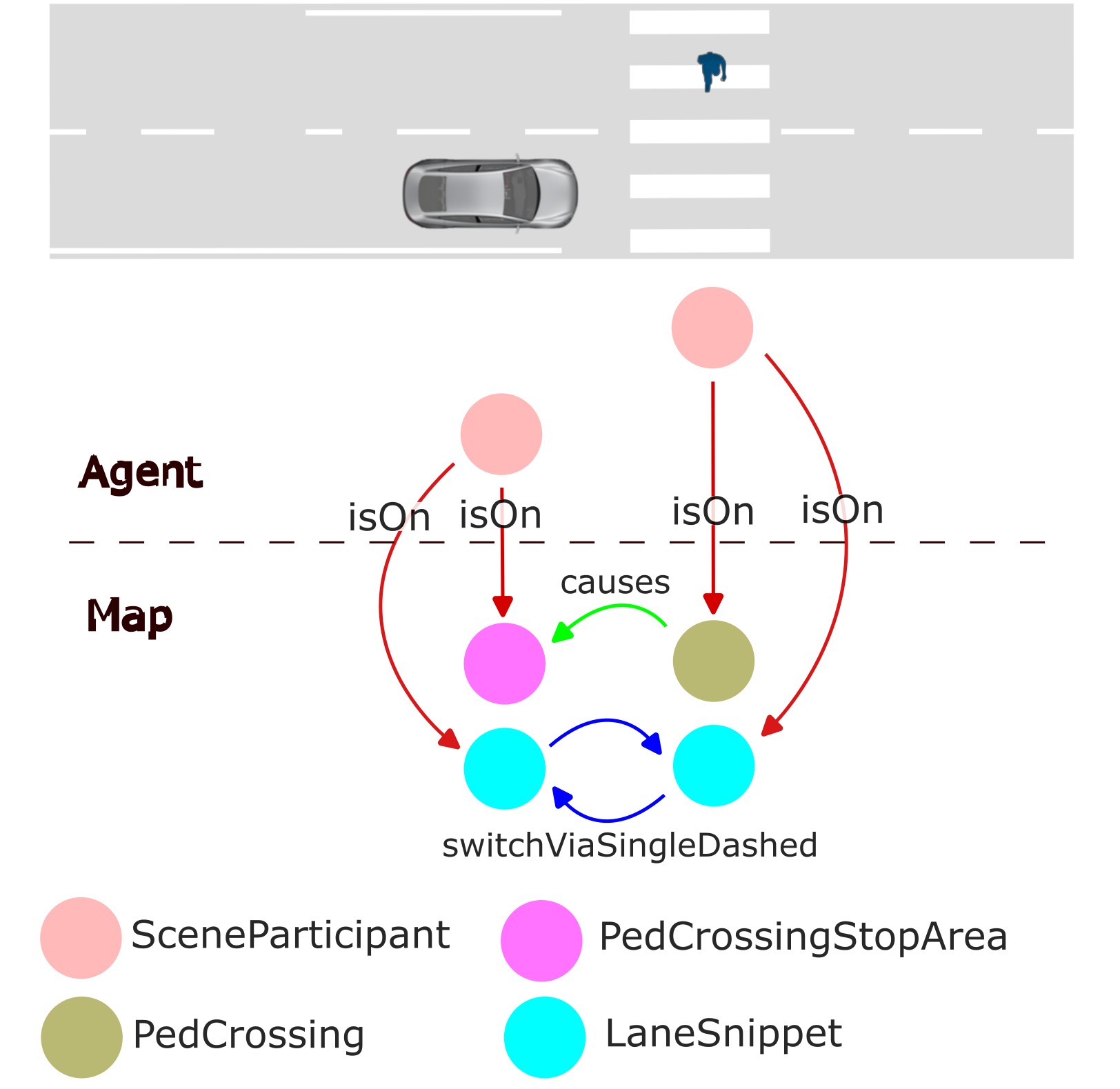}
\end{center}
   \caption{Example of how \textit{isOn} models the spatial relation between agents and map elements. In addition, the given scenario illustrates stop areas and lane snippets.}
\label{fig:long}
\label{fig:onecol}
\end{figure}

\textbf{TrafficLight.} \textit{hasTrafficLightType} differentiates horizontally and vertially stacked traffic lights. In addition, the lights are at a certain position and face a certain way, which is represented via \textit{trafficLightHasPose} pointing to a particular \textit{Pose} instance. The dynamic state of traffic lights (light color) is not modelled because this information is not available in the nuScenes dataset.
\begin{equation}
\begin{split}
    \mathit{TrafficLight} \equiv \  & \exists \mathit{trafficLightHasPose}.\mathit{Pose} \\
    \cap \  & \exists \mathit{hasTrafficLightType} . \{H, V\}
\end{split}
\end{equation}

\textit{TrafficLight} instances were generated from nuScenes \textsc{traffic\_light} records.

\textbf{PedCrossing.} This is where pedestrians can legally cross the road. The two walkways connected via a crossing are represented with the \textit{connectsWalkways} relation.
\begin{equation}
    \mathit{PedCrossing} \equiv \leq 2 \mathit{connectsWalkways} . \mathit{Walkway}
\end{equation}

Inspections of crossings and walkways in the nuScenes dataset showed that they often don't touch, but are always in close proximity. 
As a heuristic, walkways within a 5~m 
distance of a crossing were considered. 
Our algorithm chooses the two walkways with minimal distances. 
To check implementation correctness, a subset of generated triples were visualised and verified.

\textbf{Walkway, CarparkArea.} Walkways are modelled with a concept of the same name. \textit{CarparkArea} is any area where cars can park, be that on an actual carpark or by the side of a road. To represent proximity between neighbouring parts of the road explicitly, \textit{isNextTo} exists.
\begin{equation}
    \mathit{Walkway} \equiv \exists \mathit{walkwayIsNextTo}.\mathit{Lane}
\end{equation}
\begin{equation}
    \mathit{CarparkArea} \equiv \exists \mathit{carparkIsNextTo}.\mathit{Lane}
\end{equation}

The \textit{isNextTo} relation between walkways, lanes and carparks is generated for those pairs of entities that are within 4~m distance. This heuristic threshold was chosen after visualising several lanes, carparks and walkways and their proximities. 
This way an explicit spatial relation is established between neighbouring pavement surfaces.

\textbf{RoadBlock.} Road blocks group adjacent lanes that go in the same direction. A \textit{hasNextRoadBlock} edge exists from one block to another, if they contain lanes that follow one another. Road block connectivity therefore models any potential future region a car can go. Further, a \textit{hasOpposingRoadBlock} relation is introduced. It exists between two road blocks if they are parallel to each other on the same road, carrying traffic in opposite directions. This extends the spatial connectivity in the graph, making spatial relations explicit that humans see intuitively.
\begin{equation}
    \mathit{RoadBlock} \equiv \exists \mathit{hasNextRoadBlock}.\mathit{RoadBlock}
\end{equation}

\textit{RoadBlock} instances were computed by grouping neighbouring \textit{Lane}s and the connectivity between road blocks was dictated by the lane connectivity. Instances could not be generated from nuScenes \textsc{road\_block} records because they contained malformed shapes on two of the four maps, as was raised in a GitHub issue and confirmed by Motional \cite{githubIllformedMaps}. 

\textbf{Intersection.} This is where multiple lanes cross. The typical paths traversed across intersections are defined by lane connectors. A lane going into the intersection is connected to the outgoing lanes that may be travelled to legally. \textit{isConnectorOnRoadSegment} relates intersections to the lane connectors on them.
\begin{equation}
\begin{split}
    \mathit{Intersection}  \equiv \exists & \mathit{isConnectorOnRoadSegment}^{-1} . \\ & \mathit{LaneConnector}
\end{split}
\end{equation}

\textit{Intersection} instances were generated from \textsc{road\_segment} records. The explicit spatial link between them and lane connectors was computed by checking whether a lane connector overlaps with an intersection.

\textbf{hasShape.} To model the precise positions, shapes and sizes of all map elements described above, \textit{hasShape} relations are introduced for each. Each shape is represented with a subclass of the GeoSPARQL Simple Features (prefix \textit{sf}) ontology concept \textit{sf:Geometry}. 
It includes \textit{sf:Polygon}, for example, which is used to model polygonal structures like walkways, lanes or intersections. Data properties of geometries store their precise shapes in nuScenes (x, y) coordinates, but also GPS coordinates, representing the real location on Earth. This enables fusion with other geographic data sources and geospatial analysis.

\textbf{isOn, AreaElement.} To create a connection between agents and the map, the \textit{isOn} relation is introduced. \textit{AreaElement} is defined as a superclass for all map elements that occupy an area, i.e. have a \textit{sf:Polygon} geometry. \textit{isOn} links a \textit{SceneParticipant} to the map object it's currently on.
\begin{equation}
\begin{split}
    & \mathit{AreaElement} \equiv \mathit{Walkway} \cup \mathit{CarparkArea} \\
    & \quad \cup \mathit{Lane} \cup  \mathit{LaneSnippet} \cup \mathit{RoadBlock} \\ 
    & \quad \cup \mathit{StopArea} \cup \mathit{PedCrossing} \cup  \mathit{Intersection}
\end{split}
\end{equation}

The entire nSKG contains 56 million triples.

\begin{figure*}
    \centering
    \includegraphics[width=\linewidth]{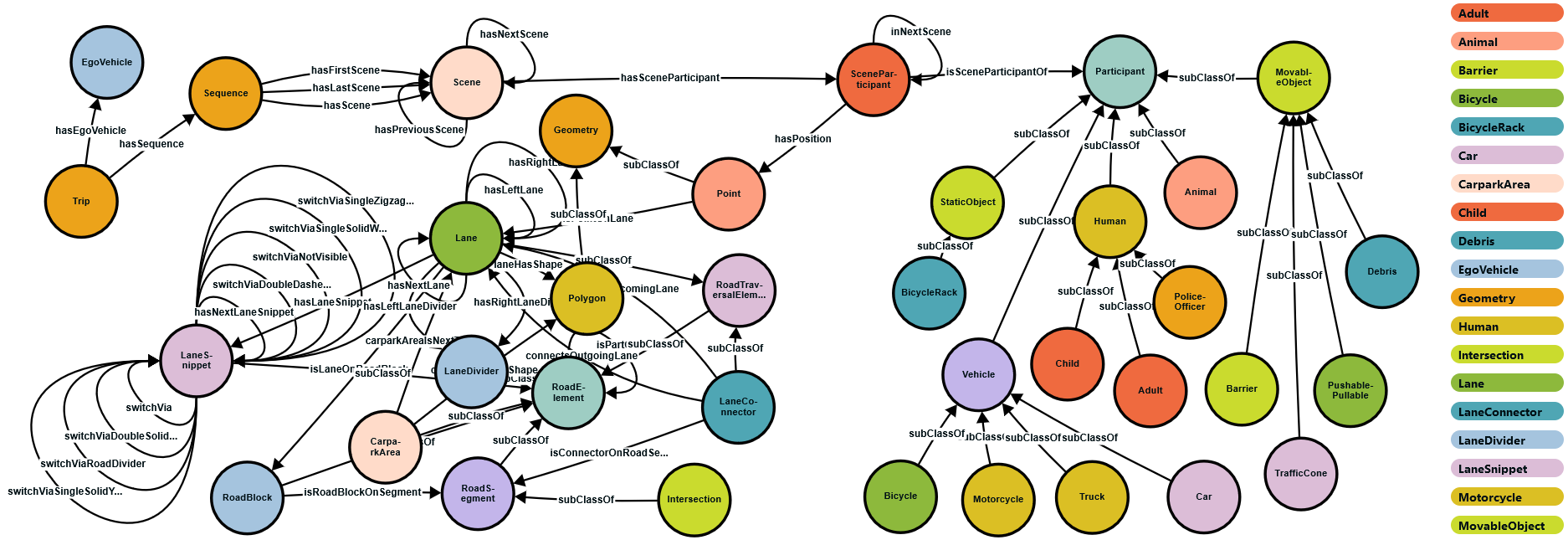}
    \caption{An excerpt of our ontology.}
    \label{fig:enter-label}
\end{figure*}


\section{nuScenes Trajectory Prediction dataset} \label{sec:dataset}

Our knowledge graph is the first resource provided in this work and contains all of the information in nuScenes as one large graph. 
It enables further research in trajectory prediction methods with information that was not readily available previously and is a step towards symbolic methods to be explored. 

However, exploring neural network models on top of our extensive representation requires a dataset of training pairs. 
nuScenes and other trajectory prediction datasets are not in this standard form and typically require extensive data preparation. 
We therefore constructed nSTP, a heterogeneous graph regression dataset for trajectory prediction. 
It comes in the format of PyTorch Geometric (PyG) \cite{Fey2019PyG}, which is one of the most widely used graph network libraries. 
The dataset is readily loadable by PyG dataloaders with input-output pairs of heterogeneous scene graphs and target trajectories.
nSTP consists of over 40,000 training pairs.

Formally, a heterogeneous graph $G = (V, E, \tau, \phi)$ has nodes $v \in V$, with node types $\tau(v)$, and edges $(u, v) \in E$, with edge types $\phi(u, v)$. The edges are directed since they are based on properties of the knowledge graph. Each example $i$ in the constructed dataset is a pair $(x_i, y_i) \in (\mathcal{G}, \mathbb{R}^{12})$, where $x_i$ is a scene graph with trajectory information from the past two seconds, local map and target identifier and $y_i$ is the ground truth future trajectory of the target. This makes our dataset a graph regression task. The constraints of 2 seconds into the past and 6 seconds into the future (sampled at 2Hz) are kept from nuScenes, such that any results on our new graph dataset can be compared to those on nuScenes raw data. The training, validation and testing splits from nuScenes are also preserved.

\subsection{Data-induced inductive bias} \label{sec:normalisation}

The coordinate system used was an important consideration as the right choice of coordinate system enables a data-centric inductive bias to be enforced, namely shift- and rotation-invariance. Inductive biases are widely considered to be essential for deep learning to generalise well~\cite{DBLP:journals/corr/abs-2011-15091,WellingMax,Bronstein2021GeometricDL}.

Coordinates in the knowledge graph (and in nuScenes) are initially in a global coordinate system. These were transformed separately for each scene graph into local, scene graph-specific coordinates, with the origin at the location of the target agent and the positive x-axis pointing along the facing direction of the target.

Precisely, let $p_{\textup{target}}$ and $R_{\textup{target}}$ be the global position (vector) and orientation (rotation matrix) of the target vehicle in scene graph $g$, respectively. Let $p_{\textup{global}}$, $R_{\textup{global}}$ be arbitrary global position and global orientation, respectively. Their representation in the local frame is given by
\begin{equation}
    p_{\textup{local}} = R_{\textup{target}}^{-1}(p_{\textup{global}} - p_{\textup{target}})
\end{equation}
\begin{equation}
    R_{\textup{local}} = R_{\textup{target}}^{-1} R_{\textup{global}}
\end{equation}
where $R_{\textup{target}}^{-1}$ is the inverse of the rotation matrix $R_{\textup{target}}$.

This way the coordinates of all entities in $g$ can be transformed into the local coordinate system. Predictions automatically become shift- and rotation-invariant because any shifts and rotations are removed in the transformation. All examples $i$ have the target at the origin, oriented along the positive x-axis. \cite{Benz2022Graph} has empirically shown that this transformation improves trajectory prediction performance.

\subsection{Participant extraction}

The trajectory information in a scene graph contains the \textit{Sequence}, \textit{Scene}, \textit{SceneParticipant} and \textit{Participant} nodes as well as the semantic relations between \textit{SceneParticipant}s. The object properties between them in the knowledge graph become heterogeneous edges. The data properties of them are turned into node features. SPARQL queries were used to retrieve the past two seconds of \textit{Scene} instances and the relevant agents in them. Relevant agents are those that may influence the target vehicle's motion, defined as those that are on a piece of relevant extracted map described next. 
This excludes, for example, scene participants on an opposing lane from consideration that have already passed the target vehicle, because they are unlikely to influence the target vehicle's future motion.

\subsection{Map extraction}

Besides trajectory information, a scene graph also contains the wealth of map information modelled in our ontology. However, including whole city maps is counter-productive and would make graphs unnecessarily large. The larger a graph, the more long-range dependencies can arise, posing problems for state-of-the-art graph neural networks \cite{Dwivedi2022LongRangeGraphBenchmark}.

Only those parts of the map were considered that affect potential paths of the target. 
To extract these from the knowledge graph, a target agent is mapped to the road block it is last on, and the \textit{hasNextRoadBlock} edges are followed four times. 
This is the maximum range travelled within 6 seconds by agents in the nuScenes training data in most cases, as our analyses showed, making this an appropriate heuristic.
Adding more into the future than necessary would make the graphs larger than necessary, hurting the performance of current graph neural networks \cite{Li2018OversmoothingPaper}. 
The map entities surrounding the potential paths are extracted via the explicit spatial relations described in the previous section. These explicit spatial relations are also kept in the heterogeneous scene graphs and, just like all the other object properties are converted into heterogeneous edges.


\section{Limitations} \label{sec:limitations}

Our ontology was tailored for representing traffic scenes in the nuScenes dataset. Despite it being a generic traffic model, it is easier to generate an associated knowledge graph from nuScenes than for other raw data sources like Argoverse.

Finally, a limitation of nSTP is that each example's $x_i$ contains between 1,000 and 2,000 nodes on average. GNNs deployed on them need to be able to handle larger graphs, which can be challenging \cite{Dwivedi2022LongRangeGraphBenchmark}.


\section{Conclusions}\label{sec:conclusions}

A comprehensive ontology for trajectory prediction has been developed with the aim to represent all relevant entities and their spatial and semantic relations in traffic scenes. 
The ontology has been tailored to the information available in the nuScenes dataset.
A knowledge graph based on the ontology has been generated from the nuScenes dataset.
The modelled concepts include many elements that were not considered previously, even by state-of-the-art approaches in trajectory prediction. 
A heterogeneous scene graph dataset was extracted from the knowledge graph, forming the first rich trajectory prediction dataset that can be immediately trained on with neural networks.
This included careful pre-processing steps to enforce rotation- and translation-invariance and to only consider agents and map elements that are relevant in each example.

The knowledge graph can be used to investigate how symbolic AI may be incorporated into trajectory prediction models. 
Reasoning with abstract entities may be a lever to increase robustness and reliability which current deep learning models lack. 
It is vital to tackle these safety issues to enable deployment of trajectory prediction algorithms in real autonomous vehicles. 
In addition, the trajectory prediction graph dataset is a major aid to future neural network research for trajectory prediction. 
It can be used to investigate novel graph neural networks that have access to richer scene information than previous approaches in trajectory prediction.

\section{Acknowledgements}

Many thanks to Benjamin Ruppik for his helpful comments and the Bosch HPC team for providing compute.

{\small
\bibliographystyle{ieee_fullname}

\begin{thebibliography}{10}\itemsep=-1pt

\bibitem{githubIllformedMaps}
{I}ll-formed maps "singapore-queenstown" and "singapore-hollandvillage" ·
  {I}ssue 862 · nutonomy/nuscenes-devkit --- github.com.
\newblock [Accessed 17-Jul-2023].

\bibitem{ASAMOpenScenario}
Asam openscenario v2.0.
\newblock Technical report, Assocation for Standardization of Automation and
  Measuring Systems, 2022.

\bibitem{ASAMOpenX}
Asam openxontology, concept.
\newblock Technical report, Assocation for Standardization of Automation and
  Measuring Systems, 2022.

\bibitem{Ahuja2022InterventionalCR}
Kartik Ahuja, Yixin Wang, Divyat Mahajan, and Yoshua Bengio.
\newblock Interventional causal representation learning.
\newblock {\em ArXiv}, abs/2209.11924, 2022.

\bibitem{Auer2007DBpediaAN}
S. Auer, Christian Bizer, Georgi Kobilarov, Jens Lehmann, Richard Cyganiak, and
  Zachary~G. Ives.
\newblock Dbpedia: A nucleus for a web of open data.
\newblock In {\em ISWC/ASWC}, 2007.

\bibitem{Bahari2021VehicleTP}
Mohammadhossein Bahari, Saeed Saadatnejad, Ahmad~Nafais Rahimi, Mohammad
  Shaverdikondori, Mohammad Shahidzadeh, Seyed-Mohsen Moosavi-Dezfooli, and
  Alexandre Alahi.
\newblock Vehicle trajectory prediction works, but not everywhere.
\newblock {\em 2022 IEEE/CVF Conference on Computer Vision and Pattern
  Recognition (CVPR)}, pages 17102--17112, 2021.

\bibitem{Battaglia2018RelationalIB}
Peter~W. Battaglia, Jessica~B. Hamrick, Victor Bapst, Alvaro Sanchez-Gonzalez,
  Vin{\'i}cius~Flores Zambaldi, Mateusz Malinowski, Andrea Tacchetti, David
  Raposo, Adam Santoro, Ryan Faulkner, Çaglar G{\"u}lçehre, H.~Francis Song,
  Andrew~J. Ballard, Justin Gilmer, George~E. Dahl, Ashish Vaswani, Kelsey~R.
  Allen, Charlie Nash, Victoria Langston, Chris Dyer, Nicolas Manfred~Otto
  Heess, Daan Wierstra, Pushmeet Kohli, Matthew~M. Botvinick, Oriol Vinyals,
  Yujia Li, and Razvan Pascanu.
\newblock Relational inductive biases, deep learning, and graph networks.
\newblock {\em ArXiv}, abs/1806.01261, 2018.

\bibitem{Benz2022Graph}
Yannik Benz.
\newblock Graph-based representations of driving scenarios for vehicle
  trajectory prediction.
\newblock Master's thesis, Technische Universitaet Darmstadt, 2022.

\bibitem{Berkemeyer2021FeasibleAA}
Hendrik Berkemeyer, Riccardo Franceschini, Tuan Tran, Lin Che, and Gordon Pipa.
\newblock Feasible and adaptive multimodal trajectory prediction with semantic
  maneuver fusion.
\newblock {\em 2021 IEEE International Conference on Robotics and Automation
  (ICRA)}, pages 8530--8536, 2021.

\bibitem{1606.01540}
Greg Brockman, Vicki Cheung, Ludwig Pettersson, Jonas Schneider, John Schulman,
  Jie Tang, and Wojciech Zaremba.
\newblock {OpenAI Gym}, 2016.

\bibitem{Bronstein2021GeometricDL}
Michael~M. Bronstein, Joan Bruna, Taco Cohen, and Petar Velivckovic.
\newblock Geometric deep learning: Grids, groups, graphs, geodesics, and
  gauges.
\newblock {\em ArXiv}, abs/2104.13478, 2021.

\bibitem{Burton2017MakingTC}
Simon Burton, Lydia Gauerhof, and Christian Heinzemann.
\newblock Making the case for safety of machine learning in highly automated
  driving.
\newblock In {\em SAFECOMP Workshops}, 2017.

\bibitem{Butz2021EventPredictiveCA}
Martin~Volker Butz, Asya Achimova, David~K. Bilkey, and Alistair Knott.
\newblock Event-predictive cognition: A root for conceptual human thought.
\newblock {\em Top. Cogn. Sci.}, 13:10--24, 2021.

\bibitem{Caesar2020nuscenes}
Holger Caesar, Varun Bankiti, Alex~H Lang, Sourabh Vora, Venice~Erin Liong,
  Qiang Xu, Anush Krishnan, Yu Pan, Giancarlo Baldan, and Oscar Beijbom.
\newblock {nuScenes: A multimodal dataset for autonomous driving}.
\newblock In {\em Proceedings of the IEEE/CVF conference on computer vision and
  pattern recognition}, pages 11621--11631, 2020.

\bibitem{Chang2019Argoverse}
Ming-Fang Chang, John Lambert, Patsorn Sangkloy, Jagjeet Singh, Sławomir Bąk,
  Andrew Hartnett, De Wang, Peter Carr, Simon Lucey, Deva Ramanan, and James
  Hays.
\newblock Argoverse: 3d tracking and forecasting with rich maps.
\newblock {\em 2019 IEEE/CVF Conference on Computer Vision and Pattern
  Recognition (CVPR)}, pages 8740--8749, 2019.

\bibitem{Clark2013WhateverNP}
Andy Clark.
\newblock Whatever next? predictive brains, situated agents, and the future of
  cognitive science.
\newblock {\em The Behavioral and brain sciences}, 36 3:181--204, 2013.

\bibitem{Cui2018MultimodalTP}
Henggang Cui, Vladan Radosavljevic, Fang-Chieh Chou, Tsung-Han Lin, Thi Nguyen,
  Tzu-Kuo Huang, Jeff~G. Schneider, and Nemanja Djuric.
\newblock Multimodal trajectory predictions for autonomous driving using deep
  convolutional networks.
\newblock {\em 2019 International Conference on Robotics and Automation
  (ICRA)}, pages 2090--2096, 2018.

\bibitem{Garcez2020NeurosymbolicAI}
Artur~S. d'Avila Garcez and L. Lamb.
\newblock Neurosymbolic ai: The 3rd wave.
\newblock {\em ArXiv}, abs/2012.05876, 2020.

\bibitem{denOuden2012HowPE}
Hanneke E.~M. den Ouden, Peter Kok, and Floris~P. de Lange.
\newblock How prediction errors shape perception, attention, and motivation.
\newblock {\em Frontiers in Psychology}, 3, 2012.

\bibitem{5206848}
Jia Deng, Wei Dong, Richard Socher, Li-Jia Li, Kai Li, and Li Fei-Fei.
\newblock Imagenet: A large-scale hierarchical image database.
\newblock In {\em 2009 IEEE Conference on Computer Vision and Pattern
  Recognition}, pages 248--255, 2009.

\bibitem{Deo2021MultimodalTP}
Nachiket Deo, Eric~M. Wolff, and Oscar Beijbom.
\newblock Multimodal trajectory prediction conditioned on lane-graph
  traversals.
\newblock In {\em Conference on Robot Learning}, 2021.

\bibitem{Djuric2018UncertaintyawareSM}
Nemanja Djuric, Vladan Radosavljevic, Henggang Cui, Thi Nguyen, Fang-Chieh
  Chou, Tsung-Han Lin, Nitin Singh, and Jeff~G. Schneider.
\newblock Uncertainty-aware short-term motion prediction of traffic actors for
  autonomous driving.
\newblock {\em 2020 IEEE Winter Conference on Applications of Computer Vision
  (WACV)}, pages 2084--2093, 2018.

\bibitem{Dwivedi2022LongRangeGraphBenchmark}
Vijay~Prakash Dwivedi, Ladislav Rampasek, Mikhail Galkin, Alipanah Parviz, Guy
  Wolf, Anh~Tuan Luu, and D. Beaini.
\newblock Long range graph benchmark.
\newblock {\em ArXiv}, abs/2206.08164, 2022.

\bibitem{Feld2011TheAO}
Michael Feld and Christian~A. M{\"u}ller.
\newblock The automotive ontology: managing knowledge inside the vehicle and
  sharing it between cars.
\newblock In {\em International Conference on Automotive User Interfaces and
  Interactive Vehicular Applications}, 2011.

\bibitem{Fey2019PyG}
Matthias Fey and Jan~E. Lenssen.
\newblock Fast graph representation learning with {PyTorch Geometric}.
\newblock In {\em ICLR Workshop on Representation Learning on Graphs and
  Manifolds}, 2019.

\bibitem{Gao2020VectorNetEH}
Jiyang Gao, Chen Sun, Hang Zhao, Yi Shen, Dragomir Anguelov, Congcong Li, and
  Cordelia Schmid.
\newblock Vectornet: Encoding hd maps and agent dynamics from vectorized
  representation.
\newblock {\em 2020 IEEE/CVF Conference on Computer Vision and Pattern
  Recognition (CVPR)}, pages 11522--11530, 2020.

\bibitem{Geyer2014ConceptAD}
Sebastian Geyer, Marcel Caspar~Attila Baltzer, Benjamin Franz, Stephan Hakuli,
  Michaela Kauer, Martin Kienle, Sonja Meier, Thomas Wei{\ss}gerber, Klaus
  Bengler, Ralph Bruder, Frank Flemisch, and Hermann Winner.
\newblock Concept and development of a unified ontology for generating test and
  use-case catalogues for assisted and automated vehicle guidance.
\newblock {\em Iet Intelligent Transport Systems}, 8:183--189, 2014.

\bibitem{DBLP:journals/corr/abs-2011-15091}
Anirudh Goyal and Yoshua Bengio.
\newblock Inductive biases for deep learning of higher-level cognition.
\newblock {\em CoRR}, abs/2011.15091, 2020.

\bibitem{Grimm2023HolisticGM}
Daniel Grimm, Philip Sch{\"o}rner, Moritz Dressler, and Johann~Marius
  Z{\"o}llner.
\newblock Holistic graph-based motion prediction.
\newblock {\em ArXiv}, abs/2301.13545, 2023.

\bibitem{Guha2015SchemaorgEO}
Ramanathan~V. Guha, Dan Brickley, and Steve Macbeth.
\newblock Schema.org: Evolution of structured data on the web.
\newblock {\em Queue}, 13:10 -- 37, 2015.

\bibitem{Halilaj2022KnowledgeGF}
Lavdim Halilaj, Juergen Luettin, Cory~Andrew Henson, and Sebastian Monka.
\newblock Knowledge graphs for automated driving.
\newblock {\em 2022 IEEE Fifth International Conference on Artificial
  Intelligence and Knowledge Engineering (AIKE)}, pages 98--105, 2022.

\bibitem{Hina2017OntologicalAM}
Manolo~Dulva Hina, Clement Thierry, Assia Soukane, and Amar Ramdane-Cherif.
\newblock Ontological and machine learning approaches for managing driving
  context in intelligent transportation.
\newblock In {\em International Conference on Knowledge Engineering and
  Ontology Development}, 2017.

\bibitem{Hogan2020KnowledgeG}
Aidan Hogan, Eva Blomqvist, Michael Cochez, Claudia d'Amato, Gerard de Melo,
  Claudio Gutierrez, Jos{\'e} Emilio~Labra Gayo, S. Kirrane, Sebastian
  Neumaier, Axel Polleres, Roberto Navigli, Axel-Cyrille~Ngonga Ngomo,
  Sabbir~M. Rashid, Anisa Rula, Lukas Schmelzeisen, Juan Sequeda, Steffen
  Staab, and Antoine Zimmermann.
\newblock Knowledge graphs.
\newblock {\em Communications of the ACM}, 64:96 -- 104, 2020.

\bibitem{Hu2013Geo-ontology}
Yingjie Hu, Krzysztof Janowicz, David Carral, Simon Scheider, Werner Kuhn, Gary
  Berg-Cross, Pascal Hitzler, Mike Dean, and Dave Kolas.
\newblock A geo-ontology design pattern for semantic trajectories.
\newblock In {\em Conference On Spatial Information Theory}, 2013.

\bibitem{ISO26262}
{Road vehicles – Functional safety}.
\newblock Standard, International Organization for Standardization, Geneva, CH,
  Mar. 2018.

\bibitem{Janowicz2018SOSAAL}
Krzysztof Janowicz, Armin Haller, Simon J.~D. Cox, Danh Le-Phuoc, and Maxime
  Lefrançois.
\newblock Sosa: A lightweight ontology for sensors, observations, samples, and
  actuators.
\newblock {\em J. Web Semant.}, 56:1--10, 2018.

\bibitem{Ji2020ASO}
Shaoxiong Ji, Shirui Pan, E. Cambria, Pekka Marttinen, and Philip~S. Yu.
\newblock A survey on knowledge graphs: Representation, acquisition, and
  applications.
\newblock {\em IEEE Transactions on Neural Networks and Learning Systems},
  33:494--514, 2020.

\bibitem{Jia2022HDGTHD}
Xiaosong Jia, Peng Wu, Li Chen, Hongyang Li, Yu~Sen Liu, and Junchi Yan.
\newblock Hdgt: Heterogeneous driving graph transformer for multi-agent
  trajectory prediction via scene encoding.
\newblock {\em ArXiv}, abs/2205.09753, 2022.

\bibitem{Klotz2018VSSoTV}
Benjamin Klotz, Raphael Troncy, Daniel Wilms, and Christian Bonnet.
\newblock Vsso: The vehicle signal and attribute ontology.
\newblock In {\em SSN@ISWC}, 2018.

\bibitem{LeCun1995Convolutional}
Yann LeCun, Yoshua Bengio, et~al.
\newblock Convolutional networks for images, speech, and time series.
\newblock {\em The handbook of brain theory and neural networks},
  3361(10):1995, 1995.

\bibitem{LeCun1998GradientbasedLA}
Yann LeCun, L{\'e}on Bottou, Yoshua Bengio, and Patrick Haffner.
\newblock Gradient-based learning applied to document recognition.
\newblock {\em Proc. IEEE}, 86:2278--2324, 1998.

\bibitem{Li2020EvolveGraphMT}
Jiachen Li, Fan Yang, Masayoshi Tomizuka, and Chiho Choi.
\newblock Evolvegraph: Multi-agent trajectory prediction with dynamic
  relational reasoning.
\newblock {\em arXiv: Computer Vision and Pattern Recognition}, 2020.

\bibitem{Li2021GRINGR}
Longyuan Li, Jinhui Yao, Li~Kevin Wenliang, Tongze He, Tianjun Xiao, Junchi
  Yan, David~Paul Wipf, and Zheng Zhang.
\newblock Grin: Generative relation and intention network for multi-agent
  trajectory prediction.
\newblock In {\em Neural Information Processing Systems}, 2021.

\bibitem{Li2018OversmoothingPaper}
Qimai Li, Zhichao Han, and Xiao-Ming Wu.
\newblock Deeper insights into graph convolutional networks for semi-supervised
  learning.
\newblock In {\em AAAI Conference on Artificial Intelligence}, 2018.

\bibitem{Li2019GRIPGI}
Xin Li, Xiaowen Ying, and Mooi~Choo Chuah.
\newblock Grip: Graph-based interaction-aware trajectory prediction.
\newblock {\em 2019 IEEE Intelligent Transportation Systems Conference (ITSC)},
  pages 3960--3966, 2019.

\bibitem{Liang2020LaneGCN}
Ming Liang, Binh Yang, Rui Hu, Yun Chen, Renjie Liao, Song Feng, and Raquel
  Urtasun.
\newblock Learning lane graph representations for motion forecasting.
\newblock {\em ArXiv}, abs/2007.13732, 2020.

\bibitem{Lin2014MicrosoftCC}
Tsung-Yi Lin, Michael Maire, Serge~J. Belongie, James Hays, Pietro Perona, Deva
  Ramanan, Piotr Doll{\'a}r, and C.~Lawrence Zitnick.
\newblock Microsoft coco: Common objects in context.
\newblock In {\em European Conference on Computer Vision}, 2014.

\bibitem{Lipton2016Mythos}
Zachary~Chase Lipton.
\newblock The mythos of model interpretability.
\newblock {\em Queue}, 16:31 -- 57, 2016.

\bibitem{liu2023laformer}
Mengmeng Liu, Hao Cheng, Lin Chen, Hellward Broszio, Jiangtao Li, Runjiang
  Zhao, Monika Sester, and Michael~Ying Yang.
\newblock Laformer: Trajectory prediction for autonomous driving with
  lane-aware scene constraints.
\newblock {\em arXiv preprint arXiv:2302.13933}, 2023.

\bibitem{Luettin2022Survey}
Juergen Luettin, Sebastian Monka, Cory~Andrew Henson, and Lavdim Halilaj.
\newblock A survey on knowledge graph-based methods for automated driving.
\newblock In {\em Iberoamerican Conference on Knowledge Graphs and Semantic
  Web}, 2022.

\bibitem{ma2019trafficpredict}
Yuexin Ma, Xinge Zhu, Sibo Zhang, Ruigang Yang, Wenping Wang, and Dinesh
  Manocha.
\newblock Trafficpredict: Trajectory prediction for heterogeneous
  traffic-agents.
\newblock In {\em Proceedings of the AAAI Conference on Artificial
  Intelligence}, volume~33, pages 6120--6127, 2019.

\bibitem{Marcus2018DeepLA}
Gary~F. Marcus.
\newblock Deep learning: A critical appraisal.
\newblock {\em ArXiv}, abs/1801.00631, 2018.

\bibitem{Monka2022Visual}
Sebastian Monka, Lavdim Halilaj, and Achim Rettinger.
\newblock A survey on visual transfer learning using knowledge graphs.
\newblock {\em ArXiv}, abs/2201.11794, 2022.

\bibitem{owl2}
B. Motik, P.~F. Patel-Schneider, and B. Parsia.
\newblock Owl 2 web ontology language: Document overview (second edition).
\newblock \url{https://www.w3.org/TR/owl2-overview/}, 2012.

\bibitem{PhanMinh2019CoverNetMB}
Tung Phan-Minh, Elena~Corina Grigore, Freddy~A. Boulton, Oscar Beijbom, and
  Eric~M. Wolff.
\newblock Covernet: Multimodal behavior prediction using trajectory sets.
\newblock {\em 2020 IEEE/CVF Conference on Computer Vision and Pattern
  Recognition (CVPR)}, pages 14062--14071, 2019.

\bibitem{Poggenhans2018Lanelet2}
Fabian Poggenhans, Jan-Hendrik Pauls, Johannes Janosovits, Stefan Orf,
  Maximilian Naumann, Florian Kuhnt, and Matthias Mayr.
\newblock Lanelet2: A high-definition map framework for the future of automated
  driving.
\newblock In {\em 2018 21st international conference on intelligent
  transportation systems (ITSC)}, pages 1672--1679. IEEE, 2018.

\bibitem{Presmeg1992TheBI}
Norma~C. Presmeg.
\newblock The body in the mind: The bodily basis of meaning, imagination and
  reason.
\newblock {\em Educational Studies in Mathematics}, 23:307--314, 1992.

\bibitem{Rajpurkar2016SQuAD1Q}
Pranav Rajpurkar, Jian Zhang, Konstantin Lopyrev, and Percy Liang.
\newblock Squad: 100,000+ questions for machine comprehension of text.
\newblock {\em ArXiv}, abs/1606.05250, 2016.

\bibitem{Sarwar2019ContextAO}
Sohail Sarwar, Saad Zia, Zia Ul-Qayyum, Muddesar Iqbal, Muhammad Safyan, Shahid
  Mumtaz, Ra{\'u}l Garc{\'i}a-Castro, and Konstantin Kostromitin.
\newblock Context aware ontology‐based hybrid intelligent framework for
  vehicle driver categorization.
\newblock {\em Transactions on Emerging Telecommunications Technologies}, 33,
  2019.

\bibitem{Scherrer2022OnTG}
Nino Scherrer, Anirudh Goyal, Stefan Bauer, Yoshua Bengio, and Nan~Rosemary Ke.
\newblock On the generalization and adaption performance of causal models.
\newblock {\em ArXiv}, abs/2206.04620, 2022.

\bibitem{Scholkopf2021TowardCR}
Bernhard Scholkopf, Francesco Locatello, Stefan Bauer, Nan~Rosemary Ke, Nal
  Kalchbrenner, Anirudh Goyal, and Yoshua Bengio.
\newblock Toward causal representation learning.
\newblock {\em Proceedings of the IEEE}, 109:612--634, 2021.

\bibitem{Summerfield2009ExpectationA}
Christopher Summerfield and Tobias Egner.
\newblock Expectation (and attention) in visual cognition.
\newblock {\em Trends in Cognitive Sciences}, 13:403--409, 2009.

\bibitem{Suryawanshi2019AnOM}
Yogita Suryawanshi, Haonan Qiu, Adel Ayara, and Birte Glimm.
\newblock An ontological model for map data in automotive systems.
\newblock {\em 2019 IEEE Second International Conference on Artificial
  Intelligence and Knowledge Engineering (AIKE)}, pages 140--147, 2019.

\bibitem{Szegedy2013IntriguingProperties}
Christian Szegedy, Wojciech Zaremba, Ilya Sutskever, Joan Bruna, D. Erhan,
  Ian~J. Goodfellow, and Rob Fergus.
\newblock Intriguing properties of neural networks.
\newblock {\em CoRR}, abs/1312.6199, 2013.

\bibitem{todorov2012mujoco}
Emanuel Todorov, Tom Erez, and Yuval Tassa.
\newblock Mujoco: A physics engine for model-based control.
\newblock In {\em 2012 IEEE/RSJ International Conference on Intelligent Robots
  and Systems}, pages 5026--5033. IEEE, 2012.

\bibitem{Ulbrich2015DefiningAS}
Simon Ulbrich, Till Menzel, Andreas Reschka, Fabian Schuldt, and Markus Maurer.
\newblock Defining and substantiating the terms scene, situation, and scenario
  for automated driving.
\newblock {\em 2015 IEEE 18th International Conference on Intelligent
  Transportation Systems}, pages 982--988, 2015.

\bibitem{Ulbrich2014GraphbasedCR}
Simon Ulbrich, Tobias Nothdurft, Markus Maurer, and Peter Hecker.
\newblock Graph-based context representation, environment modeling and
  information aggregation for automated driving.
\newblock {\em 2014 IEEE Intelligent Vehicles Symposium Proceedings}, pages
  541--547, 2014.

\bibitem{Varadarajan2021MultiPathEI}
Balakrishnan Varadarajan, Ahmed~S. Hefny, Avikalp Srivastava, Khaled~S. Refaat,
  Nigamaa Nayakanti, Andre Cornman, K.~M. Chen, Bertrand Douillard, C.~P. Lam,
  Drago Anguelov, and Benjamin Sapp.
\newblock Multipath++: Efficient information fusion and trajectory aggregation
  for behavior prediction.
\newblock {\em 2022 International Conference on Robotics and Automation
  (ICRA)}, pages 7814--7821, 2021.

\bibitem{Wang2018GLUEAM}
Alex Wang, Amanpreet Singh, Julian Michael, Felix Hill, Omer Levy, and
  Samuel~R. Bowman.
\newblock Glue: A multi-task benchmark and analysis platform for natural
  language understanding.
\newblock {\em ArXiv}, abs/1804.07461, 2018.

\bibitem{WellingMax}
Max Welling.
\newblock Do we still need models or just more data and compute?
\newblock {\em University of Amsterdam}, 2019.

\bibitem{Westhofen2022Criticality}
Lukas Westhofen, Christian Neurohr, Martin Butz, Maike Scholtes, and Michael
  Schuldes.
\newblock Using ontologies for the formalization and recognition of criticality
  for automated driving.
\newblock {\em arXiv preprint arXiv:2205.01532}, 2022.

\bibitem{Zhan2019Interaction}
Wei Zhan, Liting Sun, Di Wang, Haojie Shi, Aubrey Clausse, Maximilian Naumann,
  Julius K{\"u}mmerle, Hendrik K{\"o}nigshof, Christoph Stiller, Arnaud de
  La~Fortelle, and Masayoshi Tomizuka.
\newblock Interaction dataset: An international, adversarial and cooperative
  motion dataset in interactive driving scenarios with semantic maps.
\newblock {\em ArXiv}, abs/1910.03088, 2019.

\bibitem{Zhang2016UnderstandingDL}
Chiyuan Zhang, Samy Bengio, Moritz Hardt, Benjamin Recht, and Oriol Vinyals.
\newblock Understanding deep learning requires rethinking generalization.
\newblock {\em ArXiv}, abs/1611.03530, 2016.

\bibitem{Zhao2015CoreOF}
Lihua Zhao, Ryutaro Ichise, Seiichi Mita, and Yutaka Sasaki.
\newblock Core ontologies for safe autonomous driving.
\newblock In {\em International Workshop on the Semantic Web}, 2015.

\bibitem{Zipfl2022SceneGraphs}
Maximilian Zipfl, Felix Hertlein, Achim Rettinger, Steffen Thoma, Lavdim
  Halilaj, Juergen Luettin, Stefan Schmid, and Cory~Andrew Henson.
\newblock Relation-based motion prediction using traffic scene graphs.
\newblock {\em 2022 IEEE 25th International Conference on Intelligent
  Transportation Systems (ITSC)}, pages 825--831, 2022.

\end{thebibliography}

}

\end{document}